\documentclass[10pt,twocolumn,letterpaper]{article}

\usepackage{cvpr}
\usepackage{times}
\usepackage{epsfig}
\usepackage{graphicx}
\usepackage{amsmath}
\usepackage{amssymb}

\usepackage{subfigure}
\usepackage{amsthm}

\usepackage{multirow}
\usepackage{booktabs}
\usepackage{algorithm}
\usepackage{algorithmic}
\usepackage{epstopdf}
\usepackage{authblk}

\usepackage[pagebackref=true,breaklinks=true,letterpaper=true,colorlinks,bookmarks=false]{hyperref}

 \cvprfinalcopy 

\def\cvprPaperID{812} 

\ifcvprfinal\fi
\begin{document}

\title{Simultaneous Feature Learning and Hash Coding with Deep Neural Networks}

\author[$^{\dagger}$]{Hanjiang Lai}
\author[$^{\ddagger}$]{Yan Pan \thanks{Corresponding author: Yan Pan, email: panyan5@mail.sysu.edu.cn.}}
\author[$^\S$ ]{Ye Liu}
\author[$^{\dagger}$]{Shuicheng Yan}
\affil[$^{\dagger}$]{Department of Electronic and Computer Engineering, National University of Singapore, Singapore}
\affil[$^{\ddagger}$]{School of Software, Sun Yan-Sen University, China}
\affil[$^\S$ ]{School of Information Science and Technology, Sun Yan-Sen University, China}


\maketitle

\begin{abstract}
  Similarity-preserving hashing is a widely-used method for nearest neighbour search in large-scale image retrieval tasks. For most existing hashing methods, an image is first encoded as a vector of hand-engineering visual features, followed by another separate projection or quantization step that generates binary codes. However, such visual feature vectors may not be optimally compatible with the coding process, thus producing sub-optimal hashing codes. In this paper, we propose a deep architecture for supervised hashing, in which images are mapped into binary codes via carefully designed deep neural networks. The pipeline of the proposed deep architecture consists of three building blocks: 1) a sub-network with a stack of convolution layers to produce the effective intermediate image features; 2) a divide-and-encode module to divide the intermediate image features into multiple branches, each encoded into one hash bit; and 3) a triplet ranking loss designed to characterize that one image is more similar to the second image than to the third one. Extensive evaluations on several benchmark image datasets show that the proposed simultaneous feature learning and hash coding pipeline brings substantial improvements over other state-of-the-art supervised or unsupervised hashing methods.
\end{abstract}

\section{Introduction}
With the ever-growing large-scale image data on the Web, much
attention has been devoted to nearest neighbor search via hashing
methods. In this paper, we focus on learning-based hashing, an
emerging stream of hash methods that learn similarity-preserving
hash functions to encode input data points (e.g., images) into
binary codes.

 Many learning-based hashing methods have been proposed, e.g.,~\cite{BRE,KLSH,ITQ,KSH,MLH,CNNH,liu2013hash,wang2012semi,gong2013learning}.
The existing learning-based hashing methods can be categorized into
unsupervised and supervised methods, based on whether supervised
information (e.g., similarities or dissimilarities on data points)
is involved. Compact bitwise representations are advantageous for
improving the efficiency in both storage and search speed,
particularly in big data applications. Compared to unsupervised
methods, supervised methods usually embed the input data points into
compact hash codes with fewer bits, with the help of supervised
information.

In the pipelines of most existing hashing methods for images, each
input image is firstly represented by a vector of traditional
hand-crafted visual descriptors (e.g., GIST~\cite{GIST},
HOG~\cite{HOG}), followed by separate projection and quantization steps to
encode this vector into a binary code. However, such fixed
hand-crafted visual features  may not be optimally compatible with the coding process. In other words, a pair of
semantically similar/dissimilar images may not have feature vectors
with relatively small/large Euclidean distance. Ideally, it is expected that an image feature representation can sufficiently
preserve the image similarities, which can be learned during the
hash learning process. Very recently, Xia \textit{et
al.}~\cite{CNNH} proposed CNNH, a supervised hashing method in which
the learning process is decomposed into a stage of learning
approximate hash codes from the supervised information, followed by
a stage of simultaneously learning hash functions and image
representations based on the learned approximate hash codes.
However, in this two-stage method, the learned approximate hash
codes are used to guide the learning of the image representation,
but the learned image representation cannot give feedback for learning
better approximate hash codes. This one-way interaction thus still has limitations.

\begin{figure*}\label{overview}
  \centering
  \includegraphics[width=0.9\hsize]{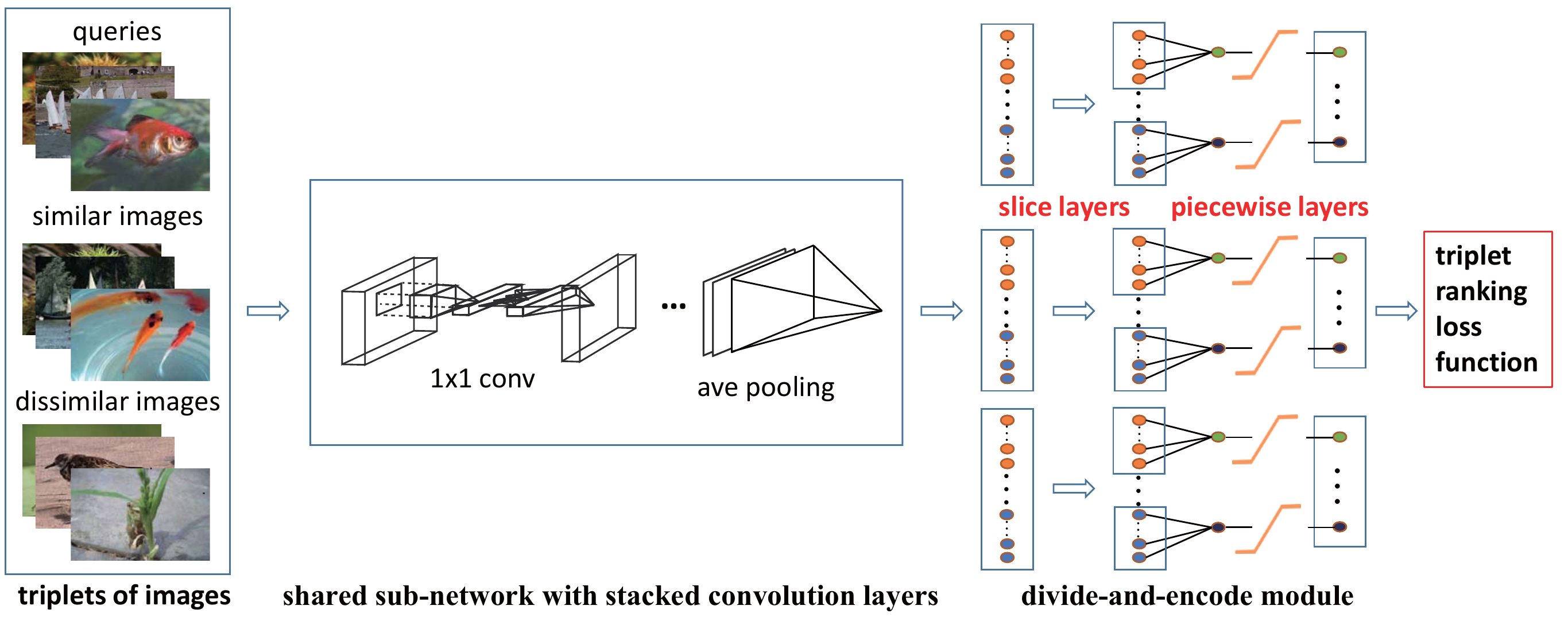}
  \caption{Overview of the proposed deep architecture for hashing. The input to the proposed architecture is in the form of triplets, i.e., $(I,I^+,I^-)$ with a query image $I$ being more similar to an image $I^+$ than to another image $I^-$. Through the proposed architecture, the image triplets are first encoded into a triplet of image feature vectors by a shared stack of multiple convolution layers. Then, each image feature vector in the triplet is converted to a hash code by a divide-and-encode module. After that, these hash codes are used in a triplet ranking loss that aims to preserve relative similarities on images.}
\end{figure*}

In this paper, we propose a ``one-stage'' supervised hashing method
via a deep architecture that maps input images to binary
codes. As shown in Figure \ref{overview}, the proposed deep
architecture has three building blocks:
 1) shared stacked
convolution layers to capture a useful image representation, 2)
divide-and-encode modules to divide intermediate image features
into multiple branches, with each branch corresponding to one hash
bit, (3) a triplet ranking loss~\cite{triplet} designed to preserve
relative similarities.
Extensive evaluations on several benchmarks
show that the proposed deep-networks-based hashing method has
substantially superior search accuracies over the state-of-the-art
supervised or unsupervised hashing methods.

\section{Related Work}
Learning-based hashing methods can be divided into two categories: unsupervised methods and supervised methods.

Unsupervised methods only use the training data to learn hash functions that can encode input data points to binary codes. Notable examples in this category include Kernelized Locality-Sensitive Hashing~\cite{KLSH}, Semantic Hashing~\cite{Semantic}, graph-based hashing methods~\cite{SH,AGH}, and Iterative Quantization~\cite{ITQ}.

Supervised methods try to leverage supervised information (e.g., class labels, pairwise similarities, or relative similarities of data points) to learn compact bitwise representations. Here are some representative examples in this category. Binary Reconstruction Embedding (BRE)~\cite{BRE} learns hash functions by minimizing the reconstruction errors between the distances of data points and those of the corresponding hash codes. Minimal Loss Hashing
(MLH)~\cite{MLH} and its extension~\cite{triplet} learn hash codes by minimizing hinge-like loss functions based on similarities or relative similarities of data points. Supervised Hashing with Kernels
(KSH)~\cite{KSH} is a kernel-based method that pursues compact binary codes to minimize the Hamming distances on similar pairs and maximize those on dissimilar pairs.

In most of the existing supervised hashing methods for images, input images are represented by some hand-crafted visual features (e.g. GIST~\cite{GIST}), before the projection and quantization steps to generate hash codes.

On the other hand, we are witnessing dramatic progress in deep convolution networks in the last few years. Approaches based on deep networks have achieved state-of-the-art performance on image classification~\cite{AlexNet,VGG,GoogleLeNet}, object detection~\cite{AlexNet,GoogleLeNet} and other recognition tasks~\cite{DeepFace}. The recent trend in convolution networks has been to increase the depth of the networks~\cite{NIN,VGG,GoogleLeNet} and the layer size~\cite{Overfeat,GoogleLeNet}. The success of deep-networks-based methods for images is mainly due to their power of automatically learning effective image representations. In this paper, we focus on a deep architecture tailored for learning-based hashing. Some parts of the proposed architecture are designed on the basis of~\cite{NIN} that uses additional $1\times 1$ convolution layers to increase the representational power of the networks.

 Without using hand-crafted image features, the recently proposed CNNH~\cite{CNNH} decomposes the hash learning process into a stage of learning approximate hash codes, followed by a deep-networks-based stage of simultaneously learning image features and hash functions, with the raw image pixels as input. However, a limitation in CNNH is that the learned image representation (in Stage 2) cannot be used to improve the learning of approximate hash codes, although the learned approximate hash codes can be used to guide the learning of image representation. In the proposed method, we learn the image representation and the hash codes in one stage, such that these two tasks have interaction and help each other forward.

\section{The Proposed Approach}
We assume $\mathcal{I}$ to be the image space. The goal of hash
learning for images is to learn a mapping
$\mathcal{F}:\mathcal{I}\rightarrow \{0,1\}^{q}$\footnote{In some of
the existing hash methods, e.g.,~\cite{CNNH,KSH}, this mapping (or
the set of hash functions) is defined as
$\mathcal{F}:\mathcal{I}\rightarrow \{-1,1\}^{q}$, which is
essentially the same as the definition used here.}, such that an
input image $I$ can be encoded into a $q$-bit binary code
$\mathcal{F}(I)$, with the similarities of images being preserved.

In this paper, we propose an architecture of deep convolution
networks designed for hash learning, as shown in Figure
\ref{overview}. This architecture accepts input images in a triplet
form. Given triplets of input images, the pipeline of the proposed
architecture contains three parts: 1) a sub-network with multiple
convolution-pooling layers to capture a representation of images; 2)
a divide-and-encode module designed to generate bitwise hash
codes; 3) a triplet ranking loss layer for learning good similarity
measures. In the following, we will present the details of these
parts, respectively.

\subsection{Triplet Ranking Loss and Optimization}
In most of the existing supervised hashing methods, the side
information is in the form of pairwise labels that indicate the
semantical similarites/dissimilarites on image pairs. The loss
functions in these methods are thus designed to preserve the
pairwise similarities of images. Recently, some
efforts~\cite{triplet,CGHash} have been made to learn hash functions
that preserve relative similarities of the form ``image $I$ is more
similar to image $I^+$ than to image $I^-$''. Such a form of
triplet-based relative similarities can be more easily obtained than
pairwise similarities (e.g., the click-through data from image
retrieval systems). Furthermore, given the side information of
pairwise similarities, one can easily generate a set of triplet
constraints\footnote{For example, for a pair of similar images
$(I_1,I_2)$ and a pair of dissimilar images $(I_1,I_3)$, one can
generate a triplet $(I_1,I_2,I_3)$ that represents ``image $I_1$ is
more similar to image $I_2$ than to image $I_3$''.}.

In the proposed deep architecture, we propose to use a variant of
the triplet ranking loss in~\cite{triplet} to preserve the relative
similarities of images. Specifically, given the training triplets of
images in the form of $(I,I^+,I^-)$ in which $I$ is more similar to
$I^+$ than to $I^-$, the goal is to find a mapping $\mathcal{F}(.)$
such that the binary code $\mathcal{F}(I)$ is closer to
$\mathcal{F}(I^+)$ than to $\mathcal{F}(I^-)$. Accordingly, the
triplet ranking hinge loss is defined by
\begin{equation}\label{Hamming_loss}
\begin{split}
&\hat{\ell}_{triplet}(\mathcal{F}(I),\mathcal{F}(I^+),\mathcal{F}(I^-))\\
=&\max(0,1-(||\mathcal{F}(I)-\mathcal{F}(I^-)||_\mathcal{H}-||\mathcal{F}(I)-\mathcal{F}(I^+)||_\mathcal{H}))\\
& s.t.\ \ \mathcal{F}(I),\ \mathcal{F}(I^+),\ \mathcal{F}(I^-)\in
\{0,1 \}^q,
\end{split}
\end{equation}
where $||.||_\mathcal{H}$ represents the Hamming distance. For ease of optimization, natural relaxation tricks on (\ref{Hamming_loss}) are to replace the Hamming norm with the $\ell_2$ norm and replace the integer constraints on $\mathcal{F}(.)$ with the range constraints. The modified loss functions is
\begin{equation}\label{loss}
\begin{split}
&\ell_{triplet}(\mathcal{F}(I),\mathcal{F}(I^+),\mathcal{F}(I^-))\\
=&\max(0,||\mathcal{F}(I)-\mathcal{F}(I^+)||_2^2-||\mathcal{F}(I)-\mathcal{F}(I^-)||_2^2+1)\\
&s.t.\ \mathcal{F}(I),\ \mathcal{F}(I^+),\ \mathcal{F}(I^-)
\in [0,1]^q.\\
\end{split}
\end{equation}
This variant of triplet ranking loss is convex. Its (sub-)gradients
with respect to $\mathcal{F}(I),\ \mathcal{F}(I^+)$ or
$\mathcal{F}(I^-)$ are
\begin{equation}
\begin{split}
&\frac{\partial{\ell}}{\partial{b}}=
(2b^--2b^+)\times I_{||b-b^+||_2^2-||b-b^-||_2^2+1>0}\\
&\frac{\partial{\ell}}{\partial{b^+}}=(2b^+-2b)\times I_{||b-b^+||_2^2-||b-b^-||_2^2+1>0}\\
&\frac{\partial{\ell}}{\partial{b^-}}=(2b^--2b)\times I_{||b-b^+||_2^2-||b-b^-||_2^2+1>0},\\
\end{split}
\end{equation}
where we denote $\mathcal{F}(I),\ \mathcal{F}(I^+)$,
$\mathcal{F}(I^-)$ as $b$, $b^+$, $b^-$. The indicator function
$I_{condition}=1$ if $condition$ is true; otherwise
$I_{condition}=0$. Hence, the loss function in (\ref{loss}) can be
easily integrated in back propagation in neural networks.

\subsection{Shared Sub-Network with Stacked Convolution Layers}
With this modified triplet ranking loss function (\ref{loss}), the input to the
proposed deep architecture are triplets of images, i.e.,
$\{(I_i,I_i^+,I_i^-)\}_{i=1}^{n}$, in which $I_i$ is more similar
to $I_i^+$ than to $I_i^-$ ($i=1,2,...n$). As shown in Figure
\ref{overview}, we propose to use a shared sub-network with a stack
of convolution layers to automatically learn a unified
representation of the input images. Through this sub-network, an
input triplet $(I,I^+,I^-)$ is encoded to a triplet of intermediate
image features $(x,x^+,x^-)$, where $x$, $x^+$,
$x^-$ are vectors with the same dimension.

In this sub-network, we adopt the architecture of Network in Network~\cite{NIN} as our
basic framework, where we insert convolution layers with $1\times 1$
filters after some convolution layers with filters of a larger
receptive field. These $1\times 1$ convolution filters can be
regarded as a linear transformation of their input channels
(followed by rectification non-linearity). As suggested in~\cite{NIN}, we use an average-pooling layer as the output layer of this sub-network, to replace the fully-connected layer(s) used in traditional architectures (e.g.,~\cite{AlexNet}). As an example, Table \ref{NIN_def} shows the configurations of the sub-network for images of size $256 \times 256$.
Note that all the convolution layers use rectification activation which are omitted in Table \ref{NIN_def}.

This sub-network is shared by the three images in each input
triplet. Such a way of parameter sharing can significantly reduce
the number of parameters in the whole architecture. A possible
alternative is that, for $(I,I^+,I^-)$ in a triplet, the query $I$
has an independent sub-network $P$, while $I^+$ and $I^-$ have a shared
sub-network $Q$, where $P$/$Q$ maps $I$/$(I^+,I^-)$ into
the corresponding image feature vector(s) (i.e., $x$, $x^+$ and
$x^-$, respectively)\footnote{Another possible alternative is that each of
$I$, $I^+$ and $I^-$ in a triplet has an independent sub-networks
(i.e., a sub-network $P$/$Q$/$R$ corresponds to $I$/$I^+$/$I^-$,
respectively), which maps it into corresponding intermediate image features.
However, such an alternative tends to get bad solutions. An extreme
example is, for any input triplets, the sub-network $P$ outputs hash
codes with all zeros; the sub-network $Q$ also outputs hash codes
with all zeros; the sub-network $R$ outputs hash codes with all
ones. Such kind of solutions may have zero loss on training data,
but their generalization performances (on test data) can be very
bad. Hence, in order to avoid such bad solutions, we consider the alternative that uses a shared sub-network for $I^+$ and $I^-$ (i.e., let $Q=R$).}. The scheme of such an alternative is similar to the idea of
``asymmetric hashing" methods~\cite{asymmetryhash}, which use two distinct hash coding
maps on a pair of images. In our experiments, we empirically show
that a shared sub-network of capturing a unified image
representation performs better than the alternative with two
independent sub-networks.


\begin{table}[t]
    \centering \caption{Configurations of the shared sub-network for input images of size $256\times 256$}
    \begin{tabular}{|c|c | c |}
        \hline
           {\bf type} &  {\bf filter size/stride} & {\bf output size} \\
        \hline
         convolution & 11$\times$ 11 / 4 & 96 $\times$ 54 $\times$ 54 \\
         \hline
         convolution &1$\times$ 1 / 1  & 96 $\times$ 54 $\times$ 54 \\
         \hline
         max pool & 3$\times$ 3 / 2  &96 $\times$ 27 $\times$ 27 \\
         \hline
         convolution & 5$\times$ 5 / 2  & 256 $\times$ 27 $\times$ 27 \\
         \hline
         convolution & 1$\times$ 1 / 1 & 256 $\times$ 27 $\times$ 27 \\
         \hline
         max pool & 3$\times$ 3 / 2 & 256 $\times$ 13 $\times$ 13 \\
         \hline
         convolution & 3$\times$ 3 / 1  & 384 $\times$ 13 $\times$ 13 \\
         \hline
         convolution & 1$\times$ 1 / 1 & 384 $\times$ 13 $\times$ 13 \\
         \hline
         max pool & 3$\times$ 3 / 2 & 384 $\times$ 6 $\times$ 6 \\
         \hline
         convolution & 3$\times$ 3 / 1   & 1024 $\times$ 6 $\times$ 6 \\
         \hline
         convolution & 1 $\times$ 1 / 1 & (50 $\times$ \# bits) $\times$ 6 $\times$ 6 \\
         \hline
         ave pool & 6$\times$ 6 / 1 &(50 $\times$ \# bits) $\times$ 1 $\times$ 1 \\
         \hline
        \end{tabular}
    \label{NIN_def}
\end{table}

\subsection{Divide-and-Encode Module}
\begin{figure}
  \centering
  \includegraphics[width=0.9\hsize]{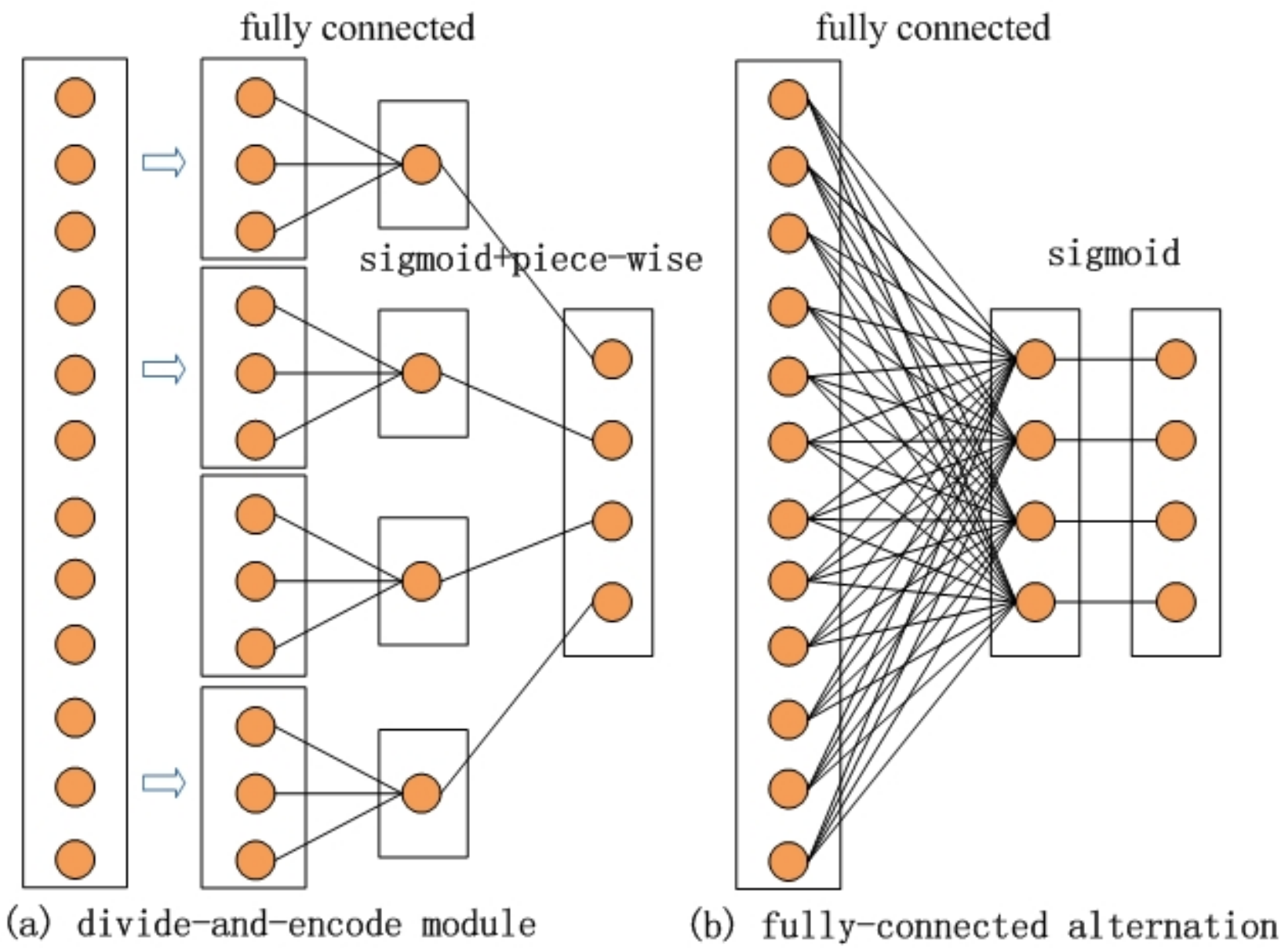}
  \caption{(a) A divide-and-encode module. (b) An alternative that consists of a fully-connected layer, followed by a sigmoid layer.}
    \label{DCM}
\end{figure}

After obtaining intermediate image features from the shared
sub-network with stacked convolution layers, we propose a
divide-and-encode module to map these image features to approximate
hash codes. We assume each target hash code has $q$ bits. Then the outputs of the shared sub-network are designed to be $50q$ (see the output size of the average-pooling layer in Table \ref{NIN_def}). As can be
seen in Figure \ref{DCM}(a), the proposed divide-and-encode
module firstly divides the input intermediate features into $q$
slices with equal length\footnote{For ease of presentation, here we
assume the dimension $d$ of the input intermediate image features is
a multiple of $q$. In practice, if $d=q\times s +c$ with $0<c<q$, we
can set the first $c$ slices to be length of $s+1$ and the rest
$q-c$ ones to be length of $s$.}. Then each slice is mapped to one
dimension by a fully-connected layer, followed by a sigmoid
activation function that restricts the output value in the range
$[0,1]$, and a piece-wise threshold function to encourage the
output of binary hash bits. After that, the $q$ output hash bits are
concatenated to be a $q$-bit (approximate) code.

As shown in Figure \ref{DCM}(b), a possible alternative to the
divide-and-encode module is a simple fully-connected layer that
maps the input intermediate image features into $q$-dimensional
vectors, followed by sigmoid activation functions to transform these
vectors into $[0,1]^q$. Compared to this alternative, the key idea
of the overall divide-and-encode strategy is trying to reduce
the redundancy among the hash bits. Specifically, in the
fully-connected alternative in Figure \ref{DCM}(b), each hash bit is
generated on the basis of the whole (and the same) input image
feature vector, which may inevitably result in redundancy among the
hash bits. On the other hand, since each hash bit is generated from
a separated slice of features, the output hash codes from the
proposed divide-and-encode module may be less redundant to each
other. Hash codes with fewer redundant bits are advocated by some
recent research. For example, the recently proposed Batch-Orthogonal
Locality Sensitive Hashing~\cite{BOLSH} theoretically and
empirically shows that hash codes generated by batch-orthogonalized
random projections are superior to those generated by simple random
projections, where batch-orthogonalized projections generate fewer
redundant hash bits than random projections. In the experiments
section, we empirically show that the proposed
divide-and-encode module leads to superior performance over the
fully-connected alternative.

In order to encourage the output of a divide-and-encode module to be binary codes, we use a sigmoid activation function followed by a piece-wise threshold function. Given a $50$-dimensional slice $x^{(i)}(i=1,2,...,q)$, the output of the $50$-to-$1$ fully-connected layer is defined by
\begin{equation}
fc_i(x^{(i)})=W_ix^{(i)},
\end{equation}
with $W_i$ being the weight matrix.

Given $c=fc_i(x^{(i)})$, the sigmoid function is defined by
\begin{equation}
sigmoid(c)=\frac{1}{1+e^{-\beta c}},
\end{equation}
where $\beta$ is a hyper-parameter.

\begin{figure}
  \centering
  \includegraphics[width=0.6\hsize]{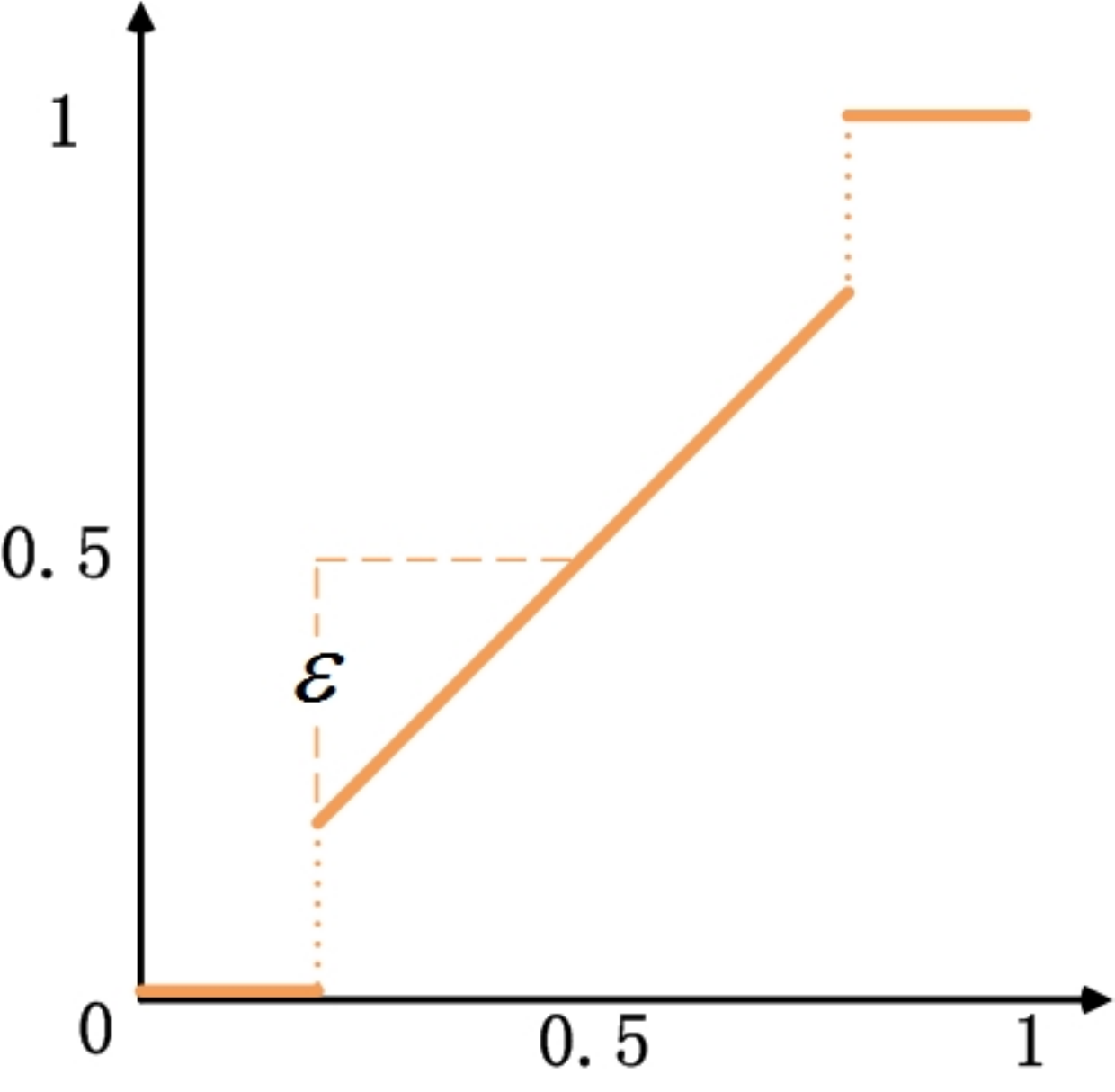}
  \caption{The piece-wise threshold function.}
    \label{piece-wise_fig}
\end{figure}

The piece-wise threshold function, as shown in Figure \ref{piece-wise_fig}, is to encourage binary outputs. Specifically, for an input variable $s=sigmoid(c)\in [0,1]$, this piece-wise function is defined by
\begin{equation}
g(s)= \left\{
\begin{array}{rl}
&0,   \quad \quad \quad \quad \quad s<0.5-\epsilon \\
&s, \quad 0.5-\epsilon \le s \le 0.5+\epsilon  \\
&1,   \quad \quad \quad \quad \quad s>0.5+\epsilon,\\
\end{array}
\right.
\end{equation}
where $\epsilon$ is a small positive hyper-parameter.

This piece-wise threshold function approximates the behavior of hard-coding, and it encourages binary outputs in training. Specifically, if the outputs from the sigmoid function are in $[0,0.5-\epsilon)$ or $(0.5+\epsilon,1]$, they are truncated to be $0$ or $1$, respectively. Note that in prediction, the proposed deep architecture only generates approximate (real-value) hash codes for input images, where these approximate codes are converted to binary codes by quantization (see Section \ref{pred} for details). With the proposed piece-wise threshold function, some of the values in the approximate hash codes (that are produced by the deep architecture) are already zeros or ones. Hence, less errors may be introduced by the quantization step.

\subsection{Hash Coding for New Images\label{pred}}
\begin{figure}[h]
  \centering
  \includegraphics[width=1.0\hsize]{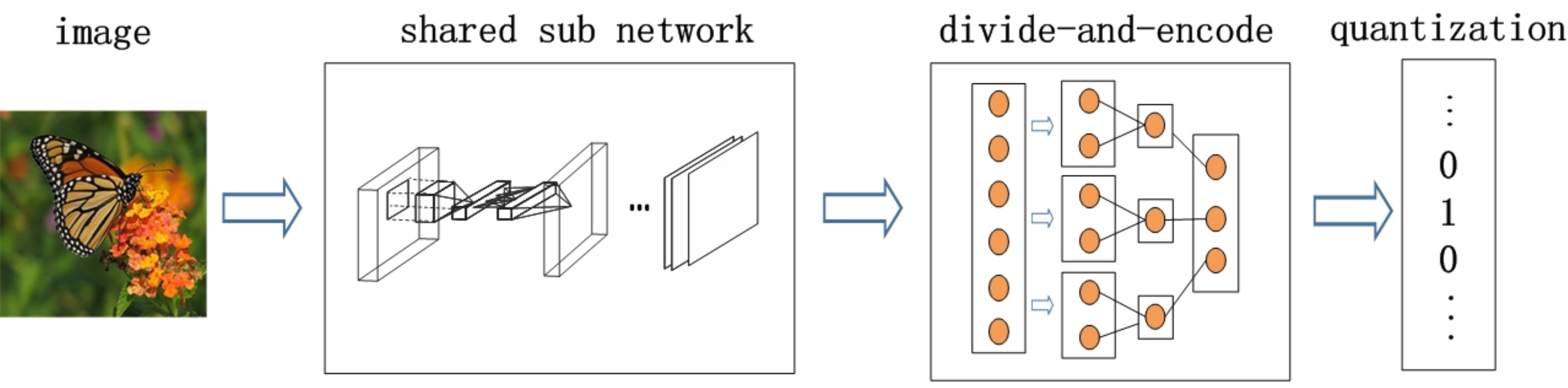}
  \caption{The architecture of prediction.}
    \label{prediction}
\end{figure}

After the deep architecture is trained, one can use it to generate a $q$-bit hash code for an input image. As shown in Figure \ref{prediction}, in prediction, an input image $I$ is first encoded into a $q$-dimensional feature vector $\mathcal{F}(I)$. Then one can obtain a $q$-bit binary code by simple quantization $b=sign(\mathcal{F}(I) - 0.5)$, where $sign(v)$ is the sign function on vectors that for $i=1,2,...,q$, $sign(v_i)=1$ if $v_i>0$, otherwise $sign(v_i)=0$.

\section{Experiments}

\begin{table*}[t]
\small
    \centering \caption{MAP of Hamming ranking w.r.t different numbers of bits on three datasets. For NUS-WIDE, we calculate the MAP values within the top 5000 returned neighbors. The results of CNNH is directly cited from~\cite{CNNH}. CNNH$\star$ is our implementation of the CNNH method in~\cite{CNNH} using Caffe, by using a network configuration comparable to that of the proposed method (see the text in Section \ref{settings} for implementation details).}
    \begin{tabular}{|c|c c c c|c c c c|c c c c|}
        \hline
\multirow{2}{*}{ Method } & \multicolumn{4}{|c}{SVHN(MAP)} &\multicolumn{4}{|c}{CIFAR-10(MAP)} & \multicolumn{4}{|c|}{NUS-WIDE(MAP)}\\
& 12 bits & 24 bits & 32 bits & 48 bits & 12 bits & 24 bits & 32 bits& 48bits & 12 bits & 24 bits &32 bits & 48 bits \\
        \hline
        Ours & {\bf 0.899 }&{\bf 0.914} & {\bf 0.925} & {\bf 0.923} & {\bf 0.552} & {\bf 0.566} & {\bf 0.558} & {\bf 0.581}  & {\bf 0.674} & {\bf 0.697} &  {\bf 0.713} &  {\bf 0.715} \\
         \hline
       CNNH$\star$ &  0.897 & 0.903 & 0.904 & 0.896 & 0.484 & 0.476  & 0.472 & 0.489 & 0.617 & 0.663 & 0.657 & 0.688 \\
         \hline
         CNNH~\cite{CNNH} & \multicolumn{4}{|c|}{N/A}& 0.439 & 0.511 & 0.509 & 0.522 & 0.611 & 0.618 & 0.625 & 0.608 \\
         \hline
         KSH~\cite{KSH} & 0.469 & 0.539 & 0.563 & 0.581 & 0.303 & 0.337 & 0.346 & 0.356 & 0.556 & 0.572 & 0.581 & 0.588  \\
         \hline
          ITQ-CCA~\cite{ITQ} & 0.428 & 0.488 & 0.489 & 0.509 & 0.264 & 0.282 & 0.288 & 0.295  & 0.435 & 0.435 & 0.435 & 0.435 \\
         \hline
          MLH~\cite{MLH} & 0.147 & 0.247& 0.261 & 0.273 & 0.182 & 0.195 & 0.207 & 0.211 & 0.500 & 0.514 & 0.520 & 0.522  \\
         \hline
          BRE~\cite{BRE}  & 0.165 & 0.206 &0.230 & 0.237 & 0.159 & 0.181 & 0.193 & 0.196   & 0.485 & 0.525 & 0.530 & 0.544\\
         \hline
         SH~\cite{SH} & 0.140 & 0.138 & 0.141 &  0.140 & 0.131 & 0.135 & 0.133 & 0.130 & 0.433 & 0.426 & 0.426 & 0.423  \\
         \hline
         ITQ~\cite{ITQ} & 0.127 & 0.132 & 0.135 & 0.139 &  0.162 & 0.169 & 0.172 & 0.175 & 0.452 & 0.468 & 0.472 & 0.477 \\
         \hline
          LSH~\cite{LSH} & 0.110 & 0.122 & 0.120 & 0.128 & 0.121 & 0.126 & 0.120 & 0.120 & 0.403 & 0.421 & 0.426 & 0.441 \\
         \hline
        \end{tabular}
    \label{map_hr}
\end{table*}

\subsection{Experimental Settings\label{settings}}
In this section, we conduct extensive evaluations of the proposed method on three benchmark datasets:
\begin{itemize}
\item
The Stree View House Numbers ({\bf SVHN})\footnote{http://ufldl.stanford.edu/housenumbers/} dataset is a real-world image dataset for recognizing digits and numbers in natural scene images. SVHN consists of over 600,000 $32\times 32$ color images in 10 classes (with digits from 0 to 9).
\item
The {\bf CIFAR-10}\footnote{http://www.cs.toronto.edu/~kriz/cifar.html} dataset consists of 60,000 color images in 10 classes. Each class has 6,000 images in size $32\times 32$.
\item
The {\bf NUS-WIDE}\footnote{http://lms.comp.nus.edu.sg/research/NUS-WIDE.htm} dataset contains nearly 270,000 images collected from Flickr.  Each of these images is associated with one or multiple labels in 81 semantic concepts. For a fair comparison, we follow the settings in~\cite{CNNH,AGH} to use the subset of images associated with the 21 most frequent labels, where each label associates with at least 5,000 images. We resize images of this subset into $256\times 256$.
\end{itemize}

We test and compare the search accuracies of the proposed method with eight state-of-the-art hashing methods, including three unsupervised methods LSH~\cite{LSH}, SH~\cite{SH} and ITQ~\cite{ITQ}, and five supervised methods CNNH~\cite{CNNH}, KSH~\cite{KSH}, MLH~\cite{MLH}, BRE~\cite{BRE} and ITQ-CCA~\cite{ITQ}.

In SVHN and CIFAR-10, we randomly select 1,000 images (100 images per class) as the test query set. For the unsupervised methods, we use the rest images as training samples. For the supervised methods, we randomly select 5,000 images (500 images per class) from the rest images as the training set. The triplets of images for training are randomly constructed based on the image class labels.

In NUS-WIDE, we randomly select 100 images from each of the selected 21 classes to form a test query set of 2,100 images. For the unsupervised methods, the rest images in the selected 21 classes are used as the training set. For supervised methods, we uniformly sample 500 images from each of the selected 21 classes to form a training set. The triplets for training are also randomly constructed based on the image class labels.

For the proposed method and CNNH, we directly use the image pixels as input. For the other baseline methods, we follow~\cite{CNNH,KSH} to represent each image in SVHN and CIFAR-10 by a 512-dimensional GIST vector; we represent each image in NUS-WIDE by a 500-dimensional bag-of-words vector \footnote{These bag-of-words features are available in the NUS-WIDE dataset.}.

To evaluate the quality of hashing, we use four evaluation metrics: Mean Average Precision ({MAP}), Precision-Recall curves, Precision curves within Hamming distance 2, and Precision curves w.r.t. different numbers of top returned samples. For a fair comparison, all of the methods use identical training and test sets.

We implement the proposed method based on the open-source \textbf{Caffe}~\cite{caffe} framework. In all experiments, our networks are trained by stochastic gradient descent with 0.9 momentum~\cite{momentum}.
We initiate $\epsilon$ in the piece-wise threshold function to be $0.5$ and decrease it by $20\%$ after every $20,000$ iterations. The mini-batch size of images is 64.
The weight decay parameter is 0.0005.

The results of BRE, ITQ, ITQ-CCA, KSH, MLH and SH are obtained by the implementations provided by their authors, respectively. The results of LSH are obtained from our implementation. Since the network configurations of CNNH in~\cite{CNNH} are different from those of the proposed method, for a fair comparison, we carefully implement CNNH (referred to as CNNH$\star$) based on Caffe, where we use the code provided by the authors of~\cite{CNNH} to implement the first stage. In the second stage of CNNH$\star$, we use the same stack of convolution-pooling layers as in Table \ref{NIN_def}, except for modifying the size of the last convolution to $bits\times 1\times 1$ and using an average pooling layer of size $bits\times 1\times 1$ as the output layer.

\subsection{Results of Search Accuracies}
Table 2 and Figure 2$\sim$4 show the comparison results of search accuracies on all of the three datasets. Two observations can be made from these results:

(1) On all of the three datasets, the proposed method achieves substantially better search accuracies (w.r.t. MAP, precision within Hamming distance 2, precision-recall, and precision with varying size of top returned samples) than those baseline methods using traditional hand-crafted visual features. For example, compared to the best competitor KSH, the MAP results of the proposed method indicate a relative increase of \textbf{58.8$\%\sim$90.6.$\%$ / 61.3$\%\sim$ 82.2 $\%$ / 21.2$\%\sim$ 22.7$\%$} on SVHN / CIFAR-10 / NUS-WIDE, respectively.

(2) In most metrics on all of the three datasets, the proposed method shows superior performance gains against the most related competitors CNNH and CNNH$\star$, which are deep-networks-based two-stage methods. For example, with respect to MAP, compared to the corresponding second best competitor, the proposed method shows a relative increase of \textbf{9.6 $\%\sim$ 14.0 $\%$ / 3.9$\%\sim$ 9.2$\%$} on CIFAR-10 / NUS-WIDE, respectively\footnote{Note that, on CIFAR-10, some MAP results of CNNH$\star$ are inferior to those of CNNH~\cite{CNNH}. This is mainly due to different network configurations and optimization frameworks between these two implementations. CNNH$\star$ is implemented based on Caffe~\cite{caffe}. But the core of the original implementation in CNNH~\cite{CNNH} is based on Cuda-Convnet~\cite{AlexNet}.}. These results verify that simultaneously learning useful representation of images and hash codes of preserving similarities can benefit each other.

\begin{figure*}[ht!]
  \begin{flushleft}
  \centering
  \subfigure[]{\label{svhn-a}
  \includegraphics[width=0.266\textwidth]{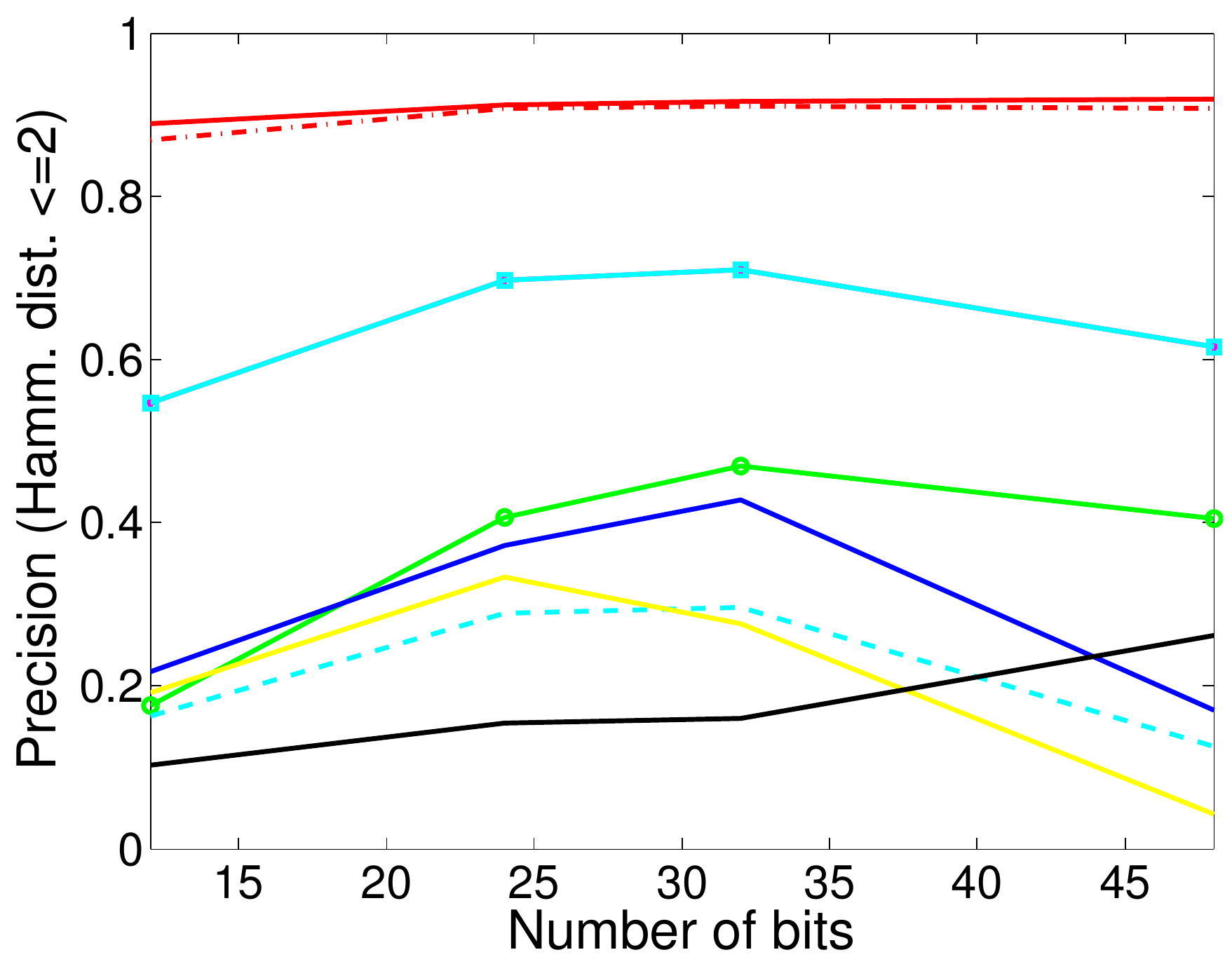}
  }
  \subfigure[]{\label{svhn-b}
  \includegraphics[width=0.266\textwidth]{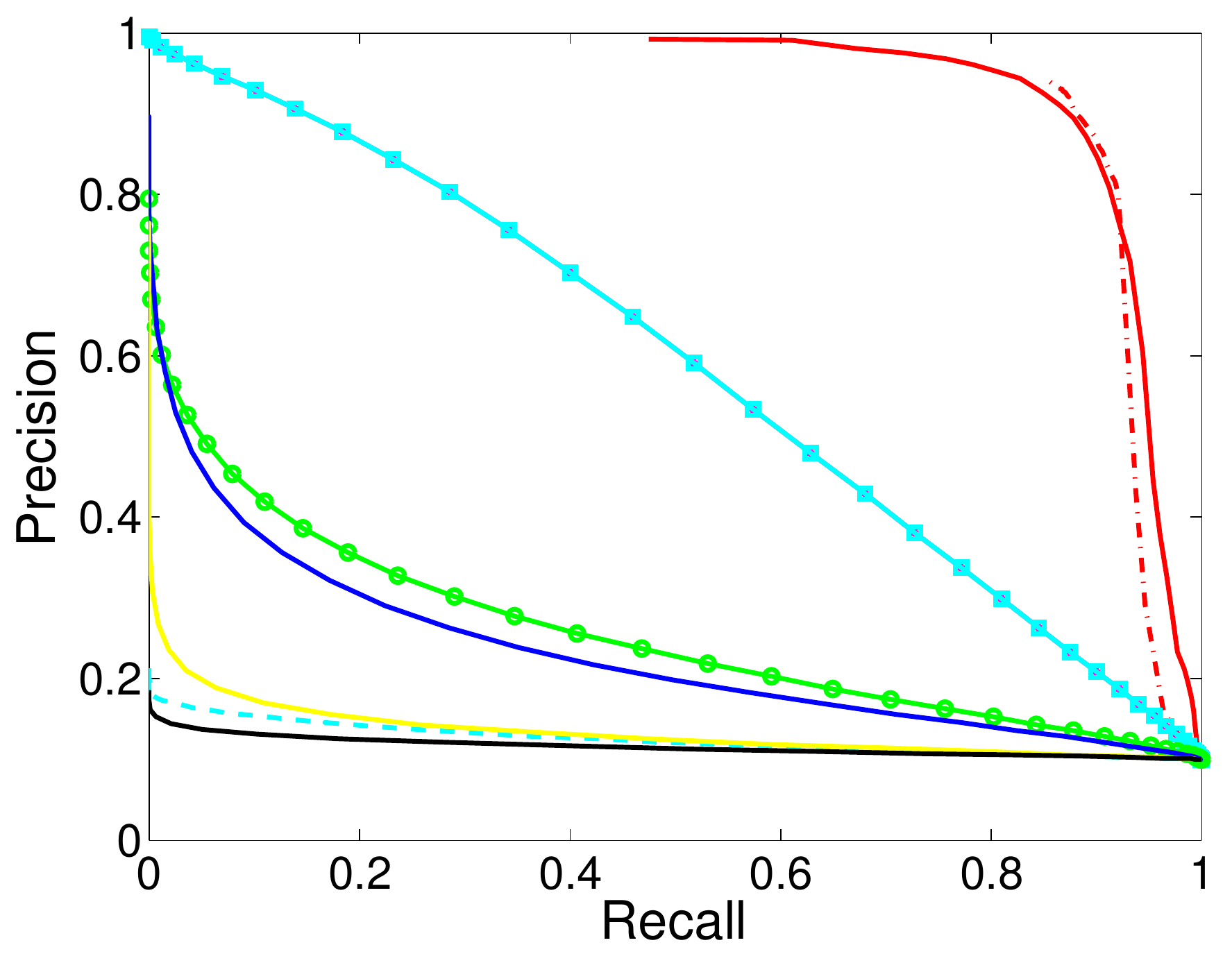}
  }
  \subfigure[]{\label{svhn-c}
  \includegraphics[width=0.327\textwidth]{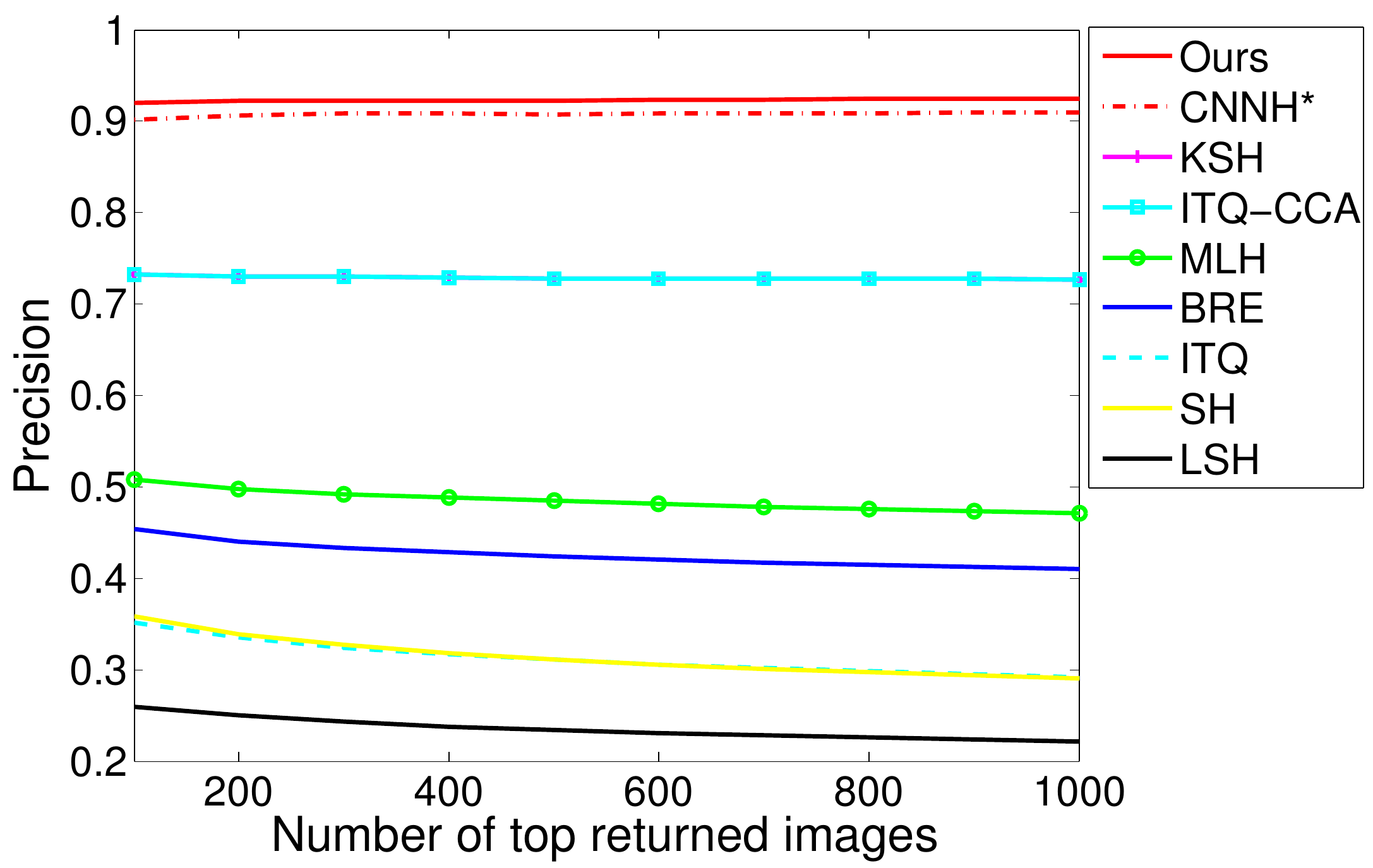}
  }
  \caption{\footnotesize The comparison results on SVNH. (a) Precision curves within Hamming radius 2; (b) precision-recall curves of Hamming ranking with 48 bits; (c) precision curves with 48 bits w.r.t. different numbers of top returned samples.}
  \label{fig: svhn-result}
  \end{flushleft}
\end{figure*}

\begin{figure*}[ht!]
  \begin{flushleft}
  \centering
  \subfigure[]{\label{cifar10-a}
  \includegraphics[width=0.266\textwidth]{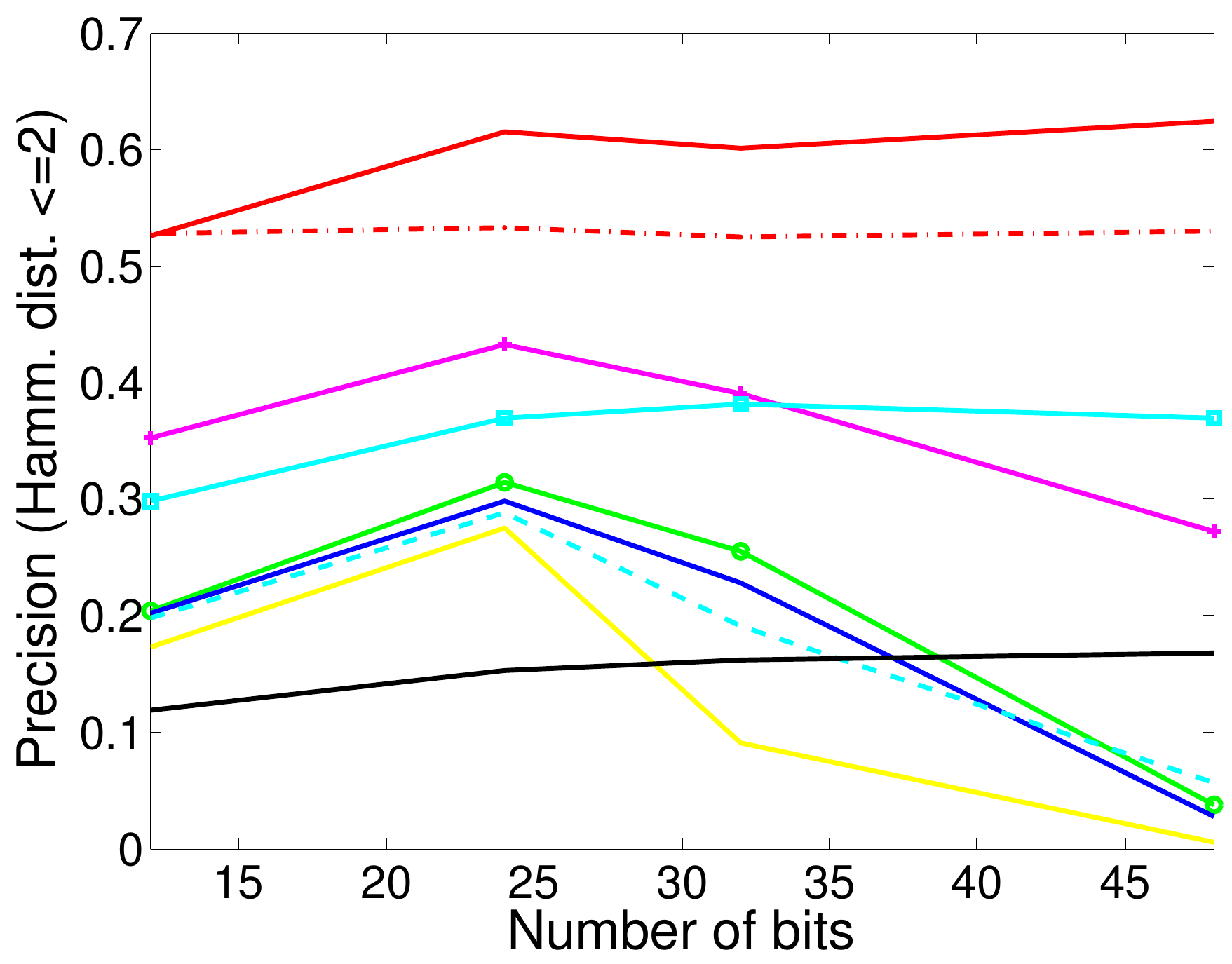}
  }
  \subfigure[]{\label{cifar10-b}
  \includegraphics[width=0.266\textwidth]{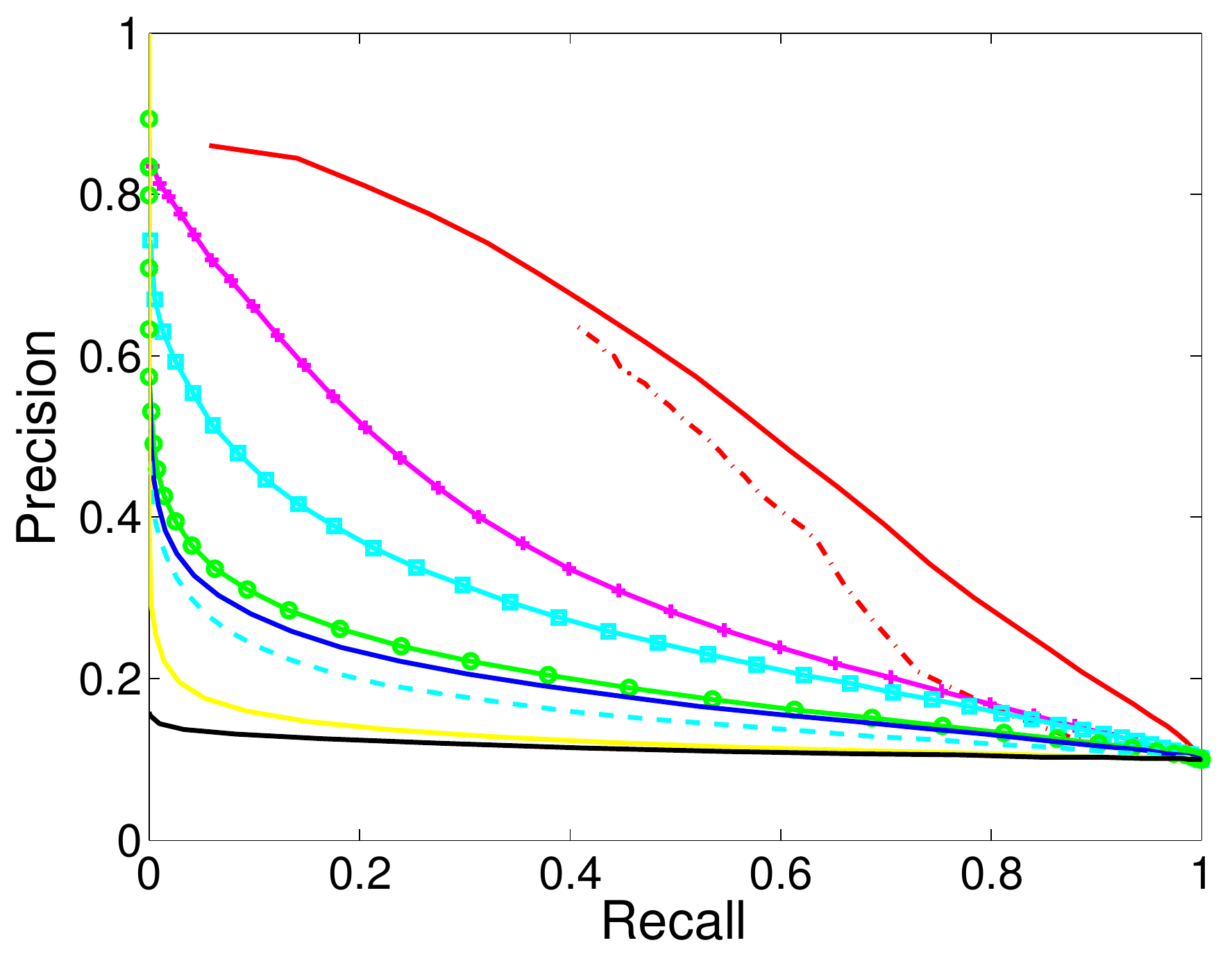}
  }
  \subfigure[]{\label{cifar10-c}
  \includegraphics[width=0.327\textwidth]{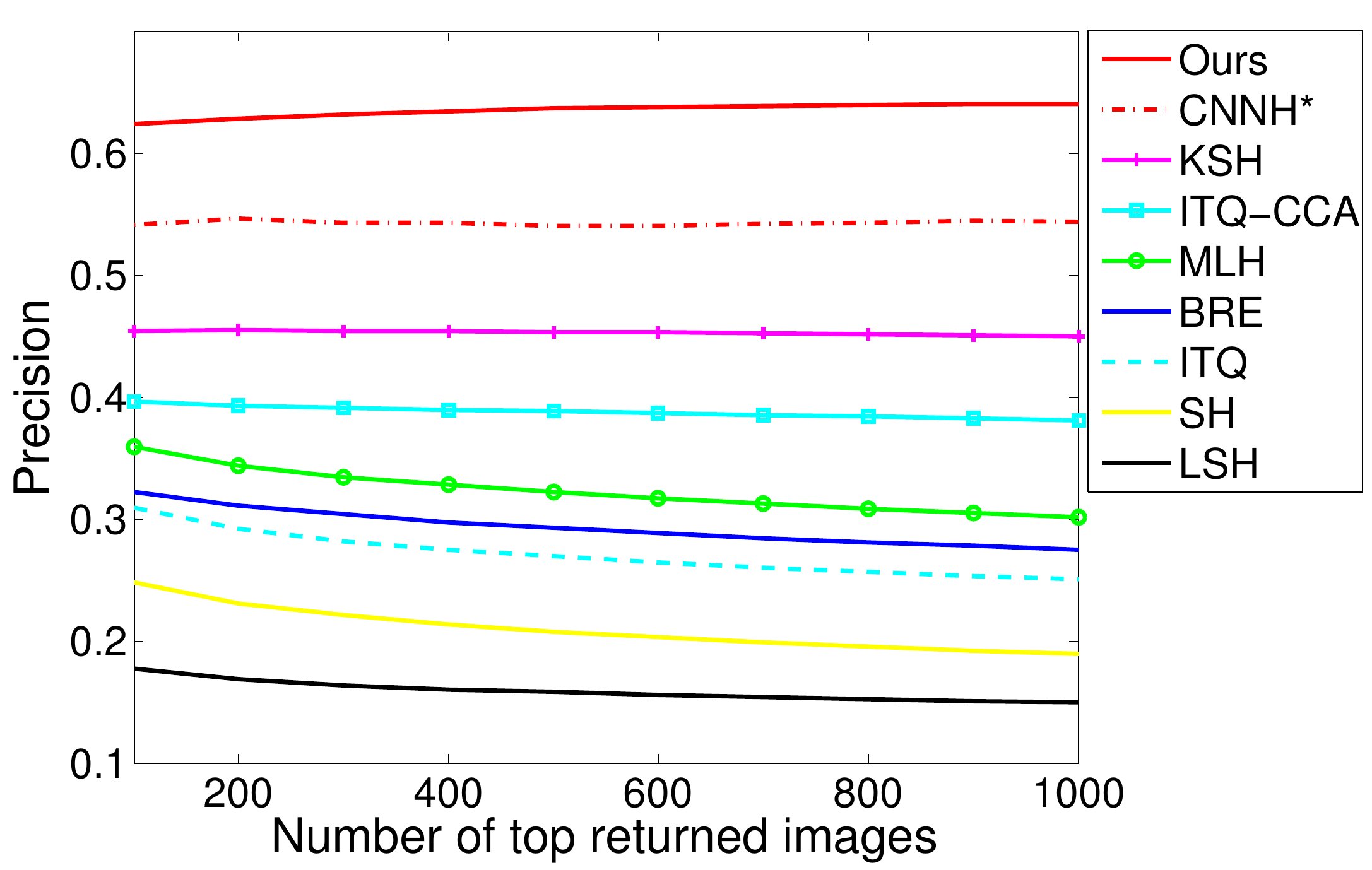}
  }
  \caption{\footnotesize The comparison results on CIFAR10. (a) precision curves within Hamming radius 2; (b) precision-recall curves of Hamming ranking with 48 bits; (c) precision curves with 48 bits w.r.t. different number of top returned samples}
  \label{fig: cifar10-result}
  \end{flushleft}
\end{figure*}
\begin{figure*}[ht!]
  \begin{flushleft}
  \centering
  \subfigure[]{\label{NUS-WIDE-a}
   \raisebox{-0.01cm}{\includegraphics[width=0.266\textwidth]{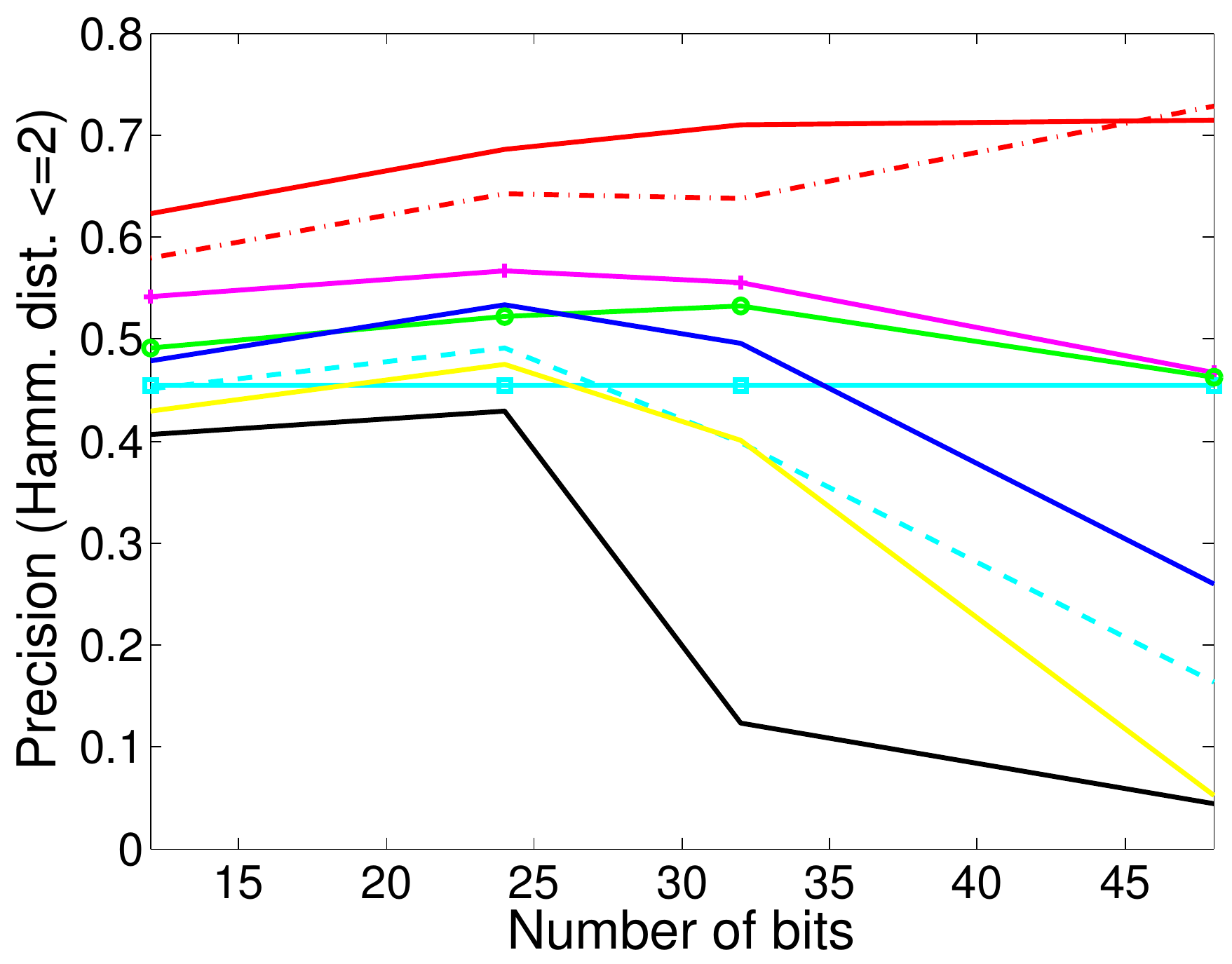}
  }}
  \subfigure[]{\label{NUS-WIDE-b}
  \raisebox{-0.01cm}{\includegraphics[width=0.266\textwidth]{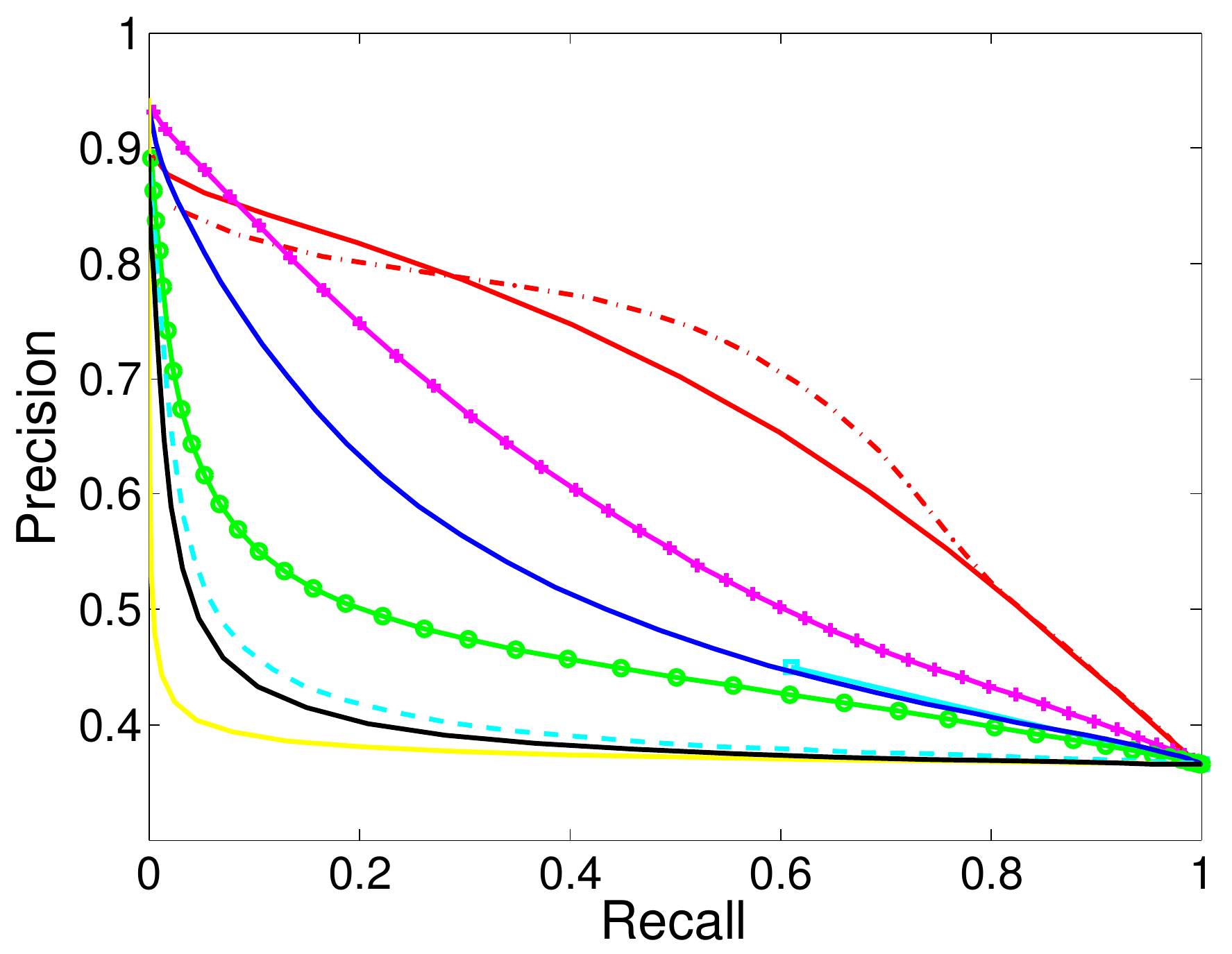}
  }}
  \subfigure[]{\label{NUS-WIDE-c}
  \raisebox{-0.01cm}{\includegraphics[width=0.327\textwidth]{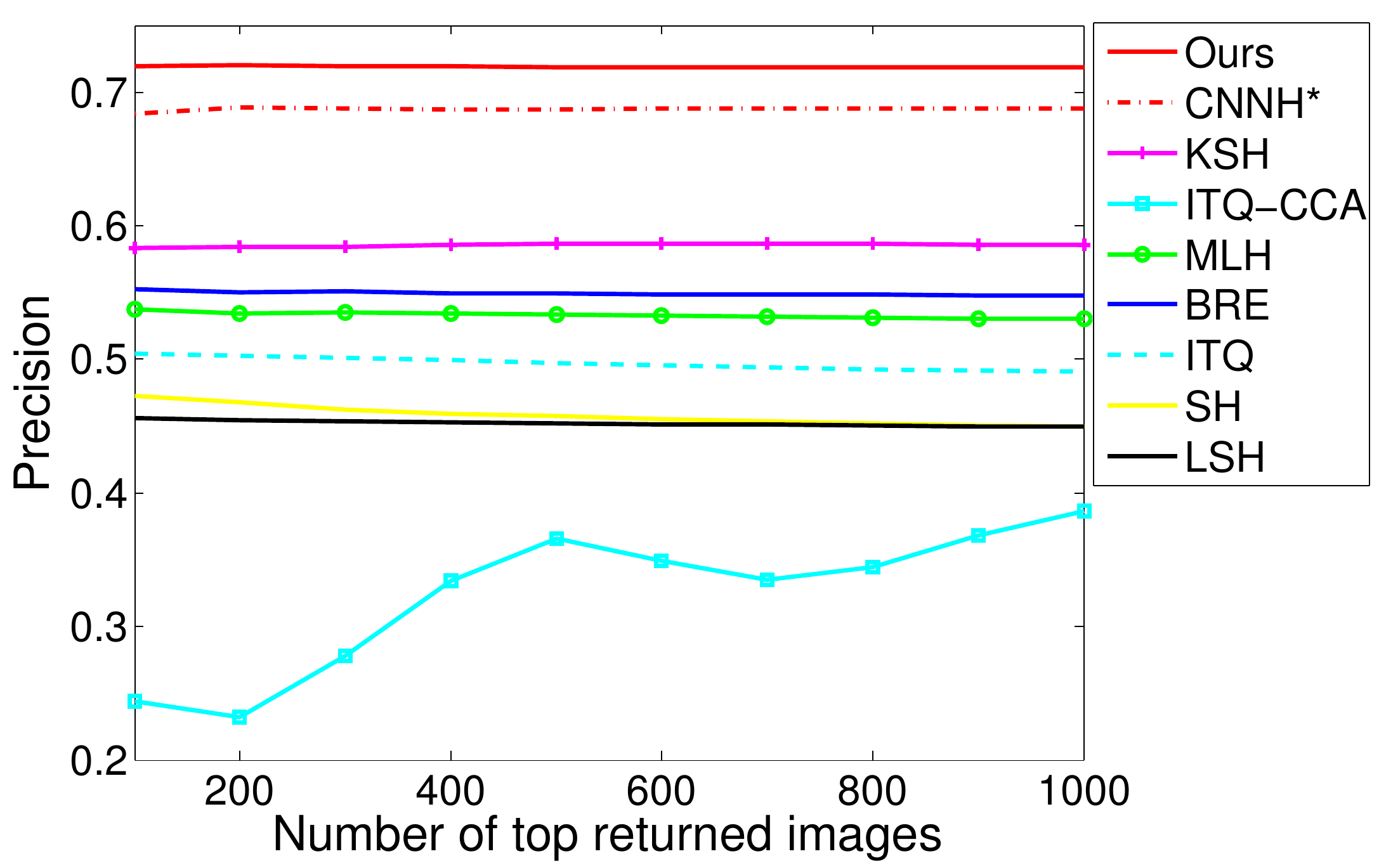}
  }}
  \caption{\footnotesize The comparison results on NUS-WIDE. (a) precision curves within Hamming radius 2; (b) precision-recall curves of Hamming ranking with 48 bits; (c) precision curves with 48 bits w.r.t. different number of top returned samples}
  \label{fig: NUSWIDE-result}
  \end{flushleft}
\end{figure*}

\subsection{Comparison Results of the Divide-and-Encode Module against Its Alternative}
A natural alternative to the divide-and-encode module is a simple fully-connected layer followed by a sigmoid layer of restricting the output values' range in $[0,1]$ (see Figure \ref{DCM}(b)). To investigate the effectiveness of the divide-and-encode module (DEM), we implement and evaluate a deep architecture derived from the proposed one in Figure \ref{overview}, by replacing the divide-and-encode module with its alternative in Figure \ref{DCM}(b) and keeping other layers unchanged. We refer to it as ``FC".

\begin{table*}[t]
\small
    \centering \caption{Comparison results of the divide-and-encode module and its fully-connected alternative on three datasets.}
    \begin{tabular}{|c|c c c c|c c c c|c c c c|}
         \hline
        \multirow{2}{*}{ Method } & \multicolumn{4}{|c}{SVHN(MAP)} &\multicolumn{4}{|c}{CIFAR-10(MAP)} & \multicolumn{4}{|c|}{NUS-WIDE(MAP)}\\
& 12 bits & 24 bits & 32 bits & 48 bits & 12 bits & 24 bits & 32 bits& 48bits & 12 bits & 24 bits &32 bits & 48 bits \\
        \hline
        Ours (DEM) & {\bf  0.899 } & {\bf 0.914} & {\bf 0.925 }&{\bf 0.923} & {\bf 0.552 }& {\bf 0.566} & {\bf 0.558} & {\bf 0.581}  & {\bf 0.674 }& {\bf 0.697} & {\bf 0.713} & {\bf 0.715}\\
         \hline
         Ours (FC) & 0.887 & 0.896 &  0.909 &0.912  & 0.465 & 0.497 & 0.489 & 0.485 & 0.623 & 0.673 & 0.682 & 0.691 \\
         \hline
        \end{tabular}
    \label{DCM-FC-MAP}
\end{table*}

\begin{figure*}[ht!]
  \begin{flushleft}
  \centering
  \subfigure[]{
  \raisebox{-0.01cm}{\includegraphics[width=0.3\textwidth]{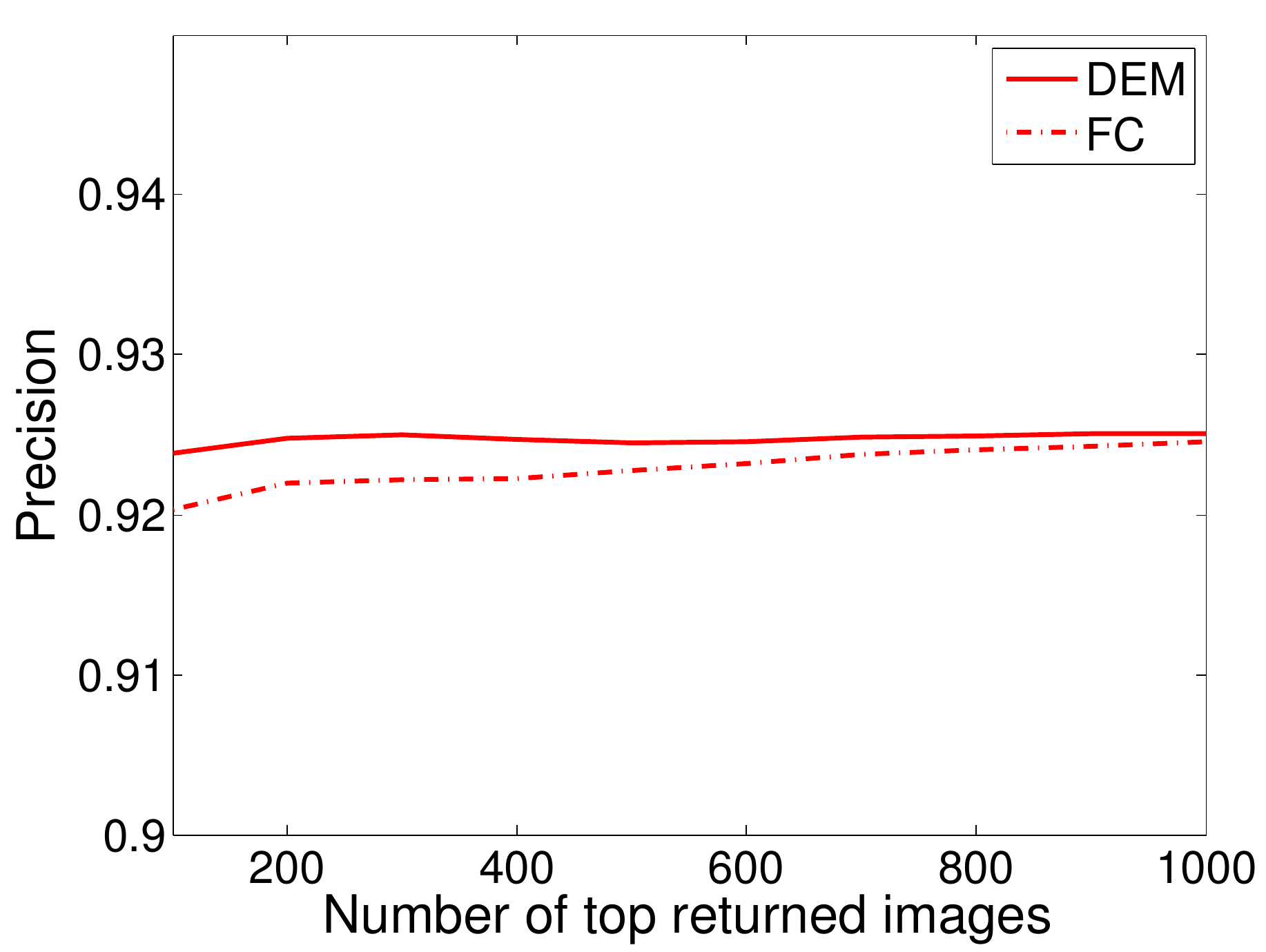}
  }}
  \subfigure[]{
  \raisebox{-0.01cm}{\includegraphics[width=0.3\textwidth]{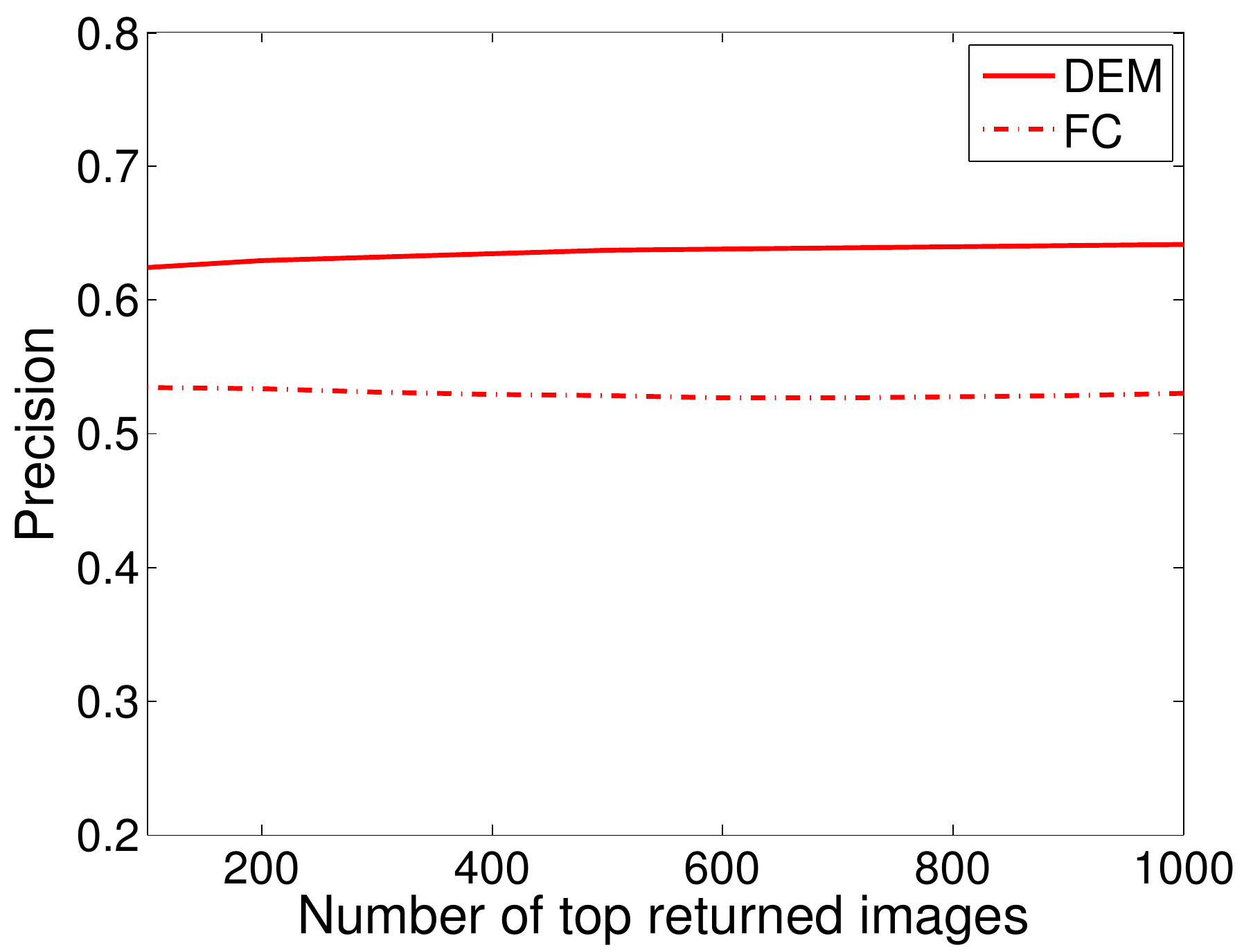}
  }}
  \subfigure[]{
  \raisebox{-0.01cm}{\includegraphics[width=0.3\textwidth]{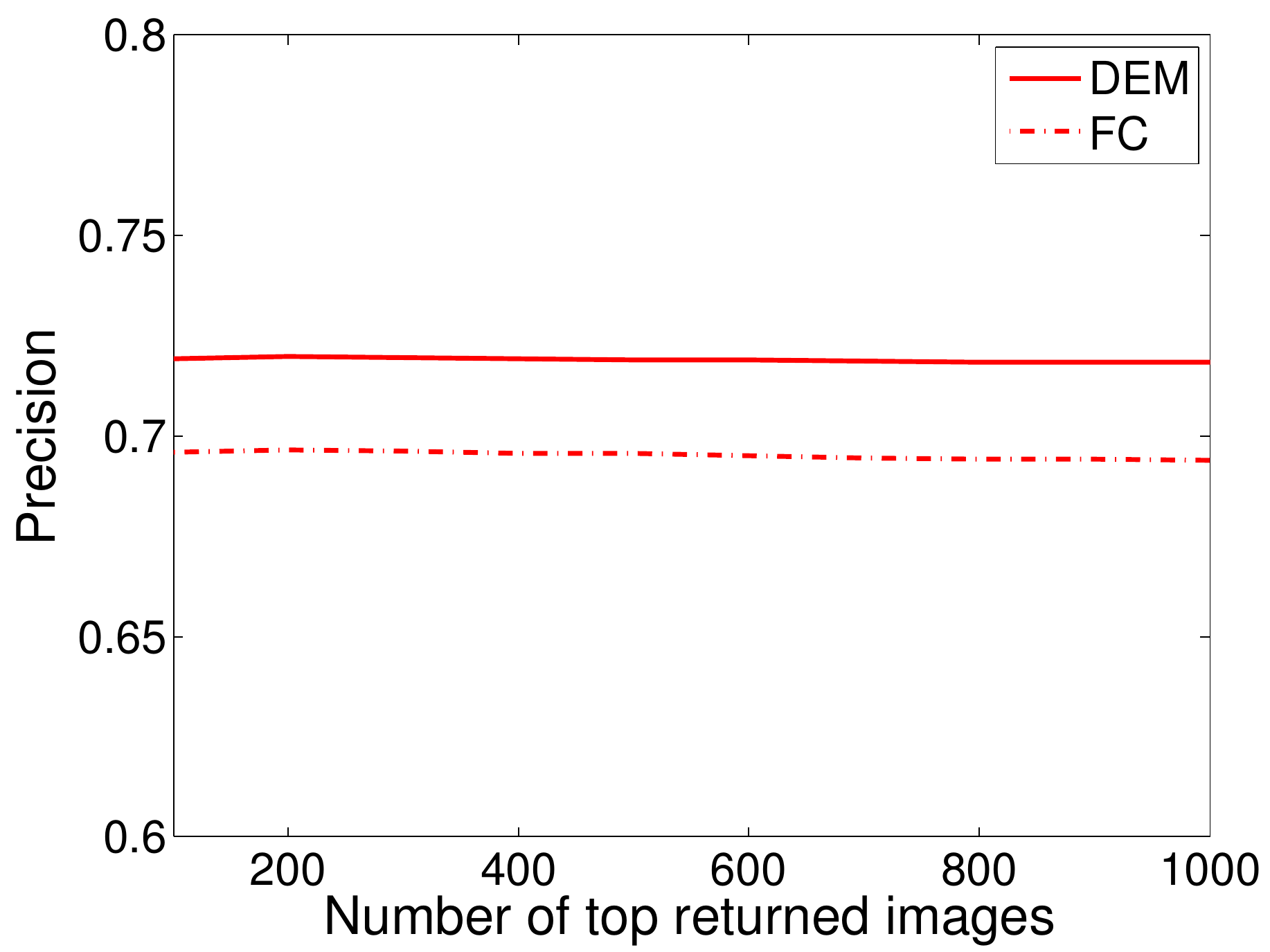}
  }}
  \caption{\footnotesize The precision curves of divide-and-encode module versus its fully-connected alternative with 48 bits w.r.t. different number of top returned samples}
  \label{DCM-FC-Prec}
  \end{flushleft}
\end{figure*}

 As can be seen from Table \ref{DCM-FC-MAP} and Figure \ref{DCM-FC-Prec}, the results of the proposed method outperform the competitor with the alternative of the divide-and-encode module. For example, the architecture with DEM achieves 0.581 accuracy with 48 bits on CIFAR-10, which indicates an improvement of 19.7$\%$ over the FC alternative. The underlying reason for the improvement may be that, compared to the FC alternative, the output hash codes from the divide-and-encode modules are less redundant to each other.

\subsection{Comparison Results of a Shared Sub-Network against Two Independent Sub-Networks}
In the proposed deep architecture, we use a shared sub-network to capture a unified image representation for the three images in an input
triplet. A possible alternative to this shared sub-network is that for a triplet $(I,I^+,I^-)$, the query $I$ has an independent sub-network $P$, while $I^+$ and $I^-$ has a shared sub-network $Q$, where $P$/$Q$ maps $I$/$(I^+,I^-)$ into the corresponding image
feature vector(s) (i.e., $x$, $x^+$ and $x^-$, respectively).

We implement and compare the search accuracies of the proposed architecture with a shared sub-network to its alternative with two independent sub-networks. As can be seen in Table \ref{sub-network-cifar} and \ref{sub-network-nus-wide}, the results of the proposed architecture outperform the competitor with the alternative with two independent sub-networks. Generally speaking, although larger networks can capture more information, it also needs more training data. The underlying reason why the architecture with a shared sub-network performs better than the one with two independent sub-networks may be that the training samples are not enough for networks with too much parameters (e.g., 500 training images per class on CIFAR-10 and NUS-WIDE).

 \begin{table}[ht!]
\small
    \centering \caption{Comparison results of a shared sub-network against two independent sub-networks on CIFAR-10.}
    \begin{tabular}{|c|c c c c|}
         \hline
Methods & 12 bits & 24 bits & 32 bits & 48 bits  \\
        \hline
        \multicolumn{5}{|c|}{MAP} \\
        \hline
        1-sub-network  & {\bf 0.552 }& {\bf 0.566} & {\bf 0.558} & {\bf 0.581}  \\
        \hline
        2-sub-networks
         &  0.467 & 0.494 & 0.477 & 0.515 \\
         \hline
         \multicolumn{5}{|c|}{Precision within Hamming radius 2} \\
        \hline
        1-sub-network  & {\bf 0.527 }& {\bf 0.615} & {\bf 0.602} & {\bf 0.625}  \\
        \hline
        2-sub-networks
         & 0.450 &  0.564 & 0.549 & 0.588 \\
         \hline
        \end{tabular}
    \label{sub-network-cifar}
\end{table}

 \begin{table}[ht!]
\small
    \centering \caption{Comparison results of a shared sub-network against two independent sub-networks on NUSWIDE.}
    \begin{tabular}{|c|c c c c|}
         \hline
Methods & 12 bits & 24 bits & 32 bits & 48 bits  \\
        \hline
        \multicolumn{5}{|c|}{MAP} \\
        \hline
        1-sub-network  & {\bf 0.674 }& {\bf 0.697} & {\bf 0.713} & {\bf 0.715}  \\
        \hline
        2-sub-networks
         & 0.640  & 0.686 & 0.688 & 0.697\\
         \hline
         \multicolumn{5}{|c|}{Precision within Hamming radius 2} \\
        \hline
          1-sub-network& {\bf 0.623 }& {\bf 0.686 } & {\bf 0.710 } & {\bf 0.714 }  \\
        \hline
        2-sub-networks
         &  0.579 & 0.664 & 0.696 & 0.704 \\
         \hline
        \end{tabular}
    \label{sub-network-nus-wide}
\end{table}

\section{Conclusion}
In this paper, we developed a ``one-stage'' supervised hashing method for image retrieval, which generates bitwise hash codes for images via a carefully designed deep architecture. The proposed deep architecture uses a triplet ranking loss designed to preserve
relative similarities. Throughout the proposed deep
architecture, input images are converted into unified image representations via a shared sub-network of stacked
convolution layers. Then, these intermediate image representations are encoded into hash codes by
divide-and-encode modules. Empirical evaluations in image retrieval show that the proposed method has superior performance gains over state-of-the-arts.

\section*{Acknowledgment}
This work was partially supported by Adobe Gift Funding. It was also  supported by the National Natural Science Foundation of China under Grants 61370021, U1401256, 61472453, Natural Science Foundation of Guangdong Province under Grant S2013010011905.


{\small
\bibliographystyle{ieee}
\bibliography{0812}
}

\end{document}